\documentclass{article}

\usepackage{microtype}

\newcommand{\spr}{\mathrm{spr}}

\usepackage{wrapfig}

\usepackage{graphicx}
\usepackage{subcaption}
\usepackage{booktabs} 
\usepackage{array}
\usepackage{xurl}      

\usepackage{hyperref}

\PassOptionsToPackage{numbers,compress}{natbib}
\usepackage[main, final]{neurips_2026}

\usepackage{amsmath}
\usepackage{amssymb}
\usepackage{mathtools}
\usepackage{amsthm}

\usepackage[capitalize,noabbrev]{cleveref}

\theoremstyle{plain}

\theoremstyle{definition}

\theoremstyle{remark}

\usepackage[textsize=tiny]{todonotes}

\title{Rank-Then-Act: Reward-Free Control from Frame-Order Progress}

\author{
    Yuriy Maksyuta 
    \And 
    George Bredis\thanks{Correspondence: \texttt{g.bredis@t-tech.dev}. Project page: \url{https://corl-team.github.io/rank-then-act}}
    \And 
    Ruslan Rakhimov 
    \And 
    Daniil Gavrilov \\ 
}

\begin{document}

\maketitle
\vspace{-35pt}
\begin{center}
T-Tech
\end{center}
\vspace{25pt}


\begin{abstract}
We introduce Rank-Then-Act (RTA), a framework for learning control policies from expert video demonstrations without environment rewards. RTA trains a Vision–Language Model (VLM) offline as a progress-based ordinal scorer, using a Group Relative Policy Optimization (GRPO) objective over shuffled frame sequences, which forces the model to recover temporal ordering from visual semantics rather than trivial time cues. Importantly, instead of using the scorer directly as a scalar reward model, we propose a correlation-based reward function for reinforcement learning: at each interaction window, we compute the Spearman rank correlation between predicted progress rankings and true temporal indices, yielding a bounded, scale-invariant learning signal. This design decouples reward learning from absolute calibration and enables stable transfer across tasks and environments. We evaluate RTA on discrete control benchmarks (PyBoy: Catrap, Kirby) and continuous control tasks (PointMaze, MetaWorld). RTA consistently matches or outperforms prior video-based reward learning methods and rank-based baselines, while demonstrating strong cross-task reuse of a single pretrained progress scorer. Our results suggest that correlation-structured supervision over video-derived ordinal signals is sufficient for policy learning, offering a scalable alternative to explicit reward design.
\end{abstract}

\section{Introduction}
Learning control from pixels \emph{without} extrinsic rewards is a foundational challenge for generalist agents, particularly in domains where reward design is brittle, unavailable, or easily exploited-such as retro games, simulators with weak task APIs, and real-world robotics, where proxy rewards often lead to specification gaming. Training agents based on vision–language models (VLMs) as reward models offers a promising path: rather than relying on hand-crafted rewards, we can derive a dense notion of \emph{progress} directly from expert video snippets and optimize policies with respect to this signal.
However, three persistent obstacles remain: (i) chronological inputs invite trivial shortcuts (\emph{later is better}), yielding vacuous, monotonically increasing progress scores; (ii) absolute progress scales are ambiguous across tasks and episodes; and (iii) online learning requires reward signals that are inexpensive to compute, sufficiently informative to guide exploration, and robust to distribution shift.
{\color{black}We propose Rank-Then-Act (RTA), a framework for reinforcement learning from video in which supervision is defined over ordinal structure rather than scalar reward prediction.

In Stage 1, we train a Vision–Language Model (VLM)~\citep{Hu2022LoRA,Qwen2.5-VL} as a progressive ordinal estimator using Group Relative Policy Optimization (GRPO)~\citep{GRPO2024} on shuffled video segments with anchor conditioning. This objective enforces recovery of temporal ordering from visual semantics while explicitly removing access to absolute frame positions, preventing trivial ``later-is-better'' solutions.

In Stage 2, we do not use the VLM as a scalar reward model. Instead, we define the reinforcement learning signal as a correlation functional over trajectories, computed as the Spearman rank correlation between predicted ordinal progress and true temporal indices within a sliding window of interaction. This produces a bounded, scale-invariant learning signal that depends only on agreement of ordering structure, rather than calibrated reward magnitude. This design enables policy learning driven entirely by progress inferred from expert demonstrations and environment interactions.}

Existing approaches only partially address this setting. Video-based reward models and VLM reward heads typically require engineered goal text, per-task tuning, or produce uncalibrated scalar values that are hard to reuse. Reward-free RL and intrinsic motivation methods assume reward queries at deployment and do not capture progress from demonstrations. Most imitation-from-observation and video IRL methods require action labels or environment rewards, which are not assumed in our setting. RTA fills this gap by providing a simple, scale-free progress signal from expert videos, using only rank correlation with time as a reward for policy learning. {\color{black}Unlike prior video reward models that directly regress scalar progress or binary success signals, RTA uses a correlation objective over ordinal structure, making the reward invariant to scale, calibration drift, and cross-task shifts.}

Our main contributions are as follows:
\begin{enumerate}
    \item We introduce a simple yet effective Vision–Language Model (VLM) \textbf{progress scorer}, trained offline on shuffled expert video clips with a GRPO objective, taught to assign \textbf{progress ranks} on shuffled clips, and kept frozen during policy learning.
    \item We propose a correlation-only reward signal for online control, computed as Spearman correlation between predicted \textbf{progress ranks} and time indices over sliding windows-enabling dense, shaped, reward-free feedback.
    {\color{black}\item We demonstrate that this reward enables effective control across discrete and continuous benchmarks without environment rewards, outperforming strong video-based baselines.
    \item We show that a single pretrained progress scorer transfers across tasks and environments, suggesting strong generality of ordinal video supervision.}
\end{enumerate}
\vspace{-2pt}

RTA contributes a simple, correlation-only reward design that enables policies to learn in fully reward-free environments from video alone, and we provide direct comparisons to prior VLM reward models and ranking-based video rewards.

\vspace{-5pt}
\section{Related Work}
\vspace{-5pt}
\textbf{Ordinal progress and ordering-based supervision from video.}
A growing line of work uses \emph{ordering} as supervision to obtain shaped rewards from passive videos.  Rank2Reward~\citep{Yang2024Rank2Reward} similarly learns shaped rewards from passive video via temporal ranking. In our experiments, we adapt Rank2Reward as a non-VLM ranking baseline using the same game video data and compare its learned rewards and downstream control performance to RTA’s correlation-only signal. VLM-RM~\citep{Rocamonde2024VLMRM} uses a clip-based progress scorer by comparing the goal completion frame with the current frame. In the original paper, it was shown that this approach often fails out-of-distribution and requires per-task environment tuning, such as changing textures or adjusting the view angle. Related efforts learn progress-sensitive embeddings or sequence rewards via temporal constraints and alignment~\citep{Zakka2022XIRL-PMLR,TimeYourRewards2024}. \emph{RTA} is closest in spirit to~\citep{Yang2024Rank2Reward} but differs in that (i) we fine-tune a compact LoRA \emph{Vision–Language Model (VLM)} once with a \emph{listwise} objective on \emph{shuffled} clips (with \emph{anchor conditioning}) to output per-frame \emph{progress ranks}, and then \emph{freeze} it; and (ii) we reward via \emph{Spearman progress–time correlation} (correlation between predicted progress ranks and timestamps) over a \emph{sliding window}, rather than learning a scalar progress head or using adversarial training.

\textbf{Vision–language models as reward/value estimators.}
Pretrained VLMs have been used as zero-shot reward models by scoring observations against natural-language goals or ordering frames to estimate value~\citep{Rocamonde2024VLMRM}, and as universal progress/value estimators by reframing value prediction as frame ordering over shuffled clips~\citep{Ma2024GVL}. VLAC~\citep{zhai2025visionlanguageactioncriticmodelroboticrealworld} train specific critic by using the same logic provided by GVL~\citep{Ma2024GVL}. These approaches leverage world knowledge in VLMs to avoid task-specific reward engineering. \emph{RTA} similarly capitalizes on a VLM prior but (i) fine-tunes it \emph{offline} as an ordinal \emph{progress scorer} on expert clips (outputting progress ranks) and (ii) converts its outputs into a correlation-based, sliding-window reward, enabling stable control from video alone without extrinsic environment rewards. We train the scorer with a \emph{GRPO-style listwise preference objective} on \emph{shuffled} clips; shuffling blocks "later-is-better" shortcuts and forces reliance on task-relevant visual cues. 

\textbf{Reward-free RL and intrinsic motivation.}
Classical reward-free RL formalizes a two-phase protocol: first explore without rewards, then solve downstream reward queries using the collected data~\citep{Jin2020RewardFree,Kaufmann2021AdaptiveRFRL,Chen2022StatEffRFRL}. In parallel, intrinsic-motivation methods (e.g., curiosity or prediction-error bonuses) provide task-agnostic signals that enable progress in sparse- or no-reward regimes~\citep{Pathak2017Curiosity}. These approaches emphasize state-space coverage or generic exploration rather than learning a task-specific progress signal from demonstrations. Our setting differs: we avoid extrinsic rewards and do not answer reward queries; instead, we optimize order consistency from demonstrations. \emph{RTA} derives a demonstration-driven \emph{ordinal} progress signal from expert video and optimizes \emph{only} a progress–time correlation objective, without coverage objectives or reward queries.

\textbf{Imitation from observation and IRL from video.}
When actions or rewards are unavailable, imitation-from-observation and IRL learn behaviors from state/video-only data. Adversarial formulations such as \emph{Generative Adversarial Imitation Learning (GAIL)} and \emph{Generative Adversarial Imitation from Observation (GAIfO)} learn policies by distribution matching without explicit reward engineering~\citep{Ho2016GAIL,Torabi2019GAIfO}. Beyond frame-wise matching, video-based methods handle temporal misalignment and embodiment gaps via sequence-level objectives and cross-video temporal constraints, including XIRL~\citep{Zakka2022XIRL-PMLR}, ordered-coverage alignment (ORCA)~\citep{Huey2025ORCA}, alignments based on \emph{Soft Dynamic Time Warping (Soft-DTW)}~\citep{TimeYourRewards2024}, and time alignment via video matching~\citep{Haresh2021LAV}. \emph{RTA} departs from adversarial or regression-style reward learning by (i) training a listwise VLM progress scorer offline, then freezing it; and (ii) rewarding online control by the \emph{Spearman} correlation between the scorer's progress ranks over a recent window and true timestamps-avoiding \emph{both} adversarial training \emph{and} scalar reward regression. Methods such as T-REX/D-REX~\citep{brown2019betterthandemonstratorimitationlearningautomaticallyranked, brown2019extrapolatingsuboptimaldemonstrationsinverse} and XIRL/ORCA~\citep{Zakka2022XIRL-PMLR, Huey2025ORCA} operate in a different data regime: they require multiple demonstration trajectories of the same task or decompose the task into subgoals, whereas our setting assumes only raw videos with no actions or rewards. Adapting them would require either recovering actions from human play or changing the problem formulation. We therefore focus our empirical comparison on video-only reward models and ranking baselines (Rank2Reward, VLM-RM) that can be applied without action labels, and leave adapting action-dependent IRL methods to video-only VLMs for future work. VICtoR~\citep{hung2025victorlearninghierarchicalvisioninstruction} is also not directly comparable to our setting, as it is a vision–instruction correlation method that relies on natural-language task descriptions and additional supervision (e.g., motion-level annotations and object/state signals).

\vspace{-7pt}

\section{Method}
\label{sec:method}
\vspace{-5pt}

We present Rank-Then-Act (RTA), a two-stage framework that converts a model’s grasp of frame order in expert videos into a bounded learning signal in $[-1,1]$ for reward-free control. Stage~1 (Rank): fine-tune a VLM as a listwise progress scorer that assigns each frame a progress rank (larger = later), trained with GRPO~\citep{GRPO2024} using Spearman’s rank correlation between predicted scores and ground-truth timestamps as the reward. Stage~2 (Act): freeze the scorer and, during interaction, at query steps compute Spearman’s $\rho$ over a sliding window between the scorer’s outputs and the window’s timestamps; this correlation is the sole reward for policy-gradient training of the agent (no extrinsic environment rewards). We next formalize the control setting and define this correlation-based reward primitive.

\vspace{-7pt}

\subsection{Problem Setting and Correlation Primitive}
\label{sec:setup}

\vspace{-5pt}

We consider goal-conditioned, partially observed Markov decision processes with observation space $\mathcal{S}$, action space $\mathcal{A}$, transition function $\mathcal{P}$, episode horizon $\mathcal{T}$, and a textual goal $g$. The goal $g$ is supplied as part of the agent’s textual prompt to initialize the task. We do not, however, condition on $g$ in the conventional RL sense; $g$ is simply included in the prompt for goal-directed tasks. The agent $\pi_\theta: \mathcal{S} \to \Delta(\mathcal{A})$ maximizes expected return using an implicit, model-derived reward. When extrinsic rewards are absent, learning relies solely on an estimated measure of \emph{progress consistency} produced by a model trained on expert video demonstrations.

Our sole scalar signal in both stages is the Spearman rank correlation,
\[
\spr(x, y) \;\coloneqq\; \mathrm{Pearson}\!\big(\mathrm{rank}(x),\, \mathrm{rank}(y)\big) \in [-1, 1],
\]
We use $\spr(\cdot, \cdot)$ both as the bounded, task-agnostic training objective in Stage~1 and as the reward signal in Stage~2. The idea of using rank correlation follows~\citep{Ma2024GVL} as an estimator of trajectory correctness.

\begin{figure*}[t]
    \centering
    \includegraphics[width=\linewidth]{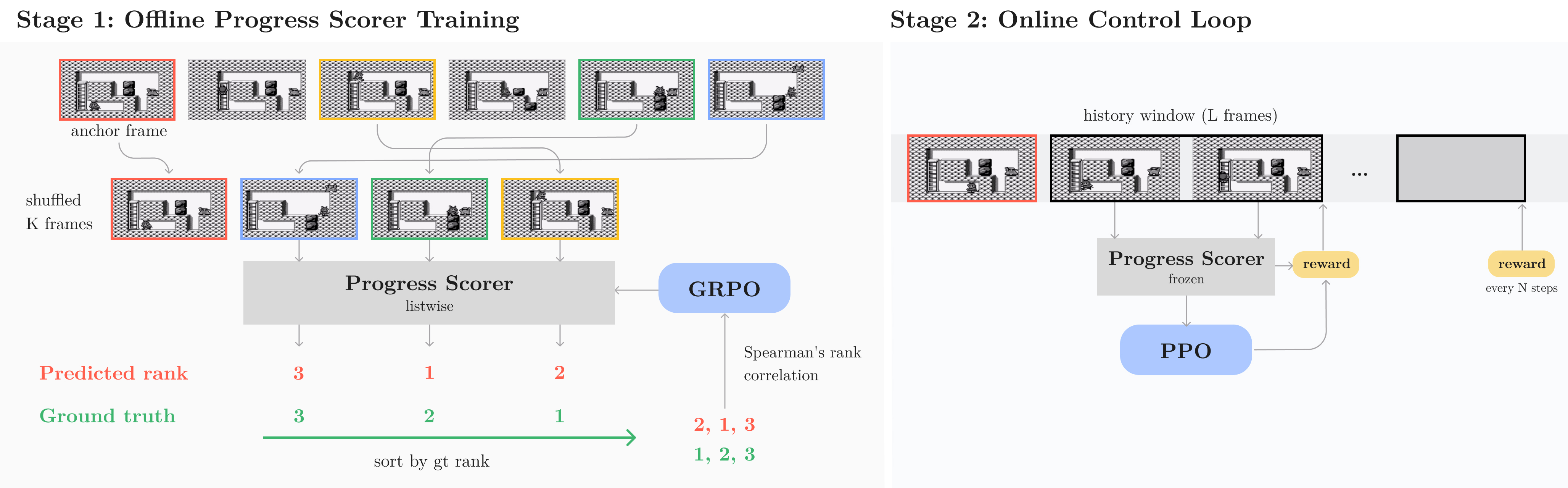}
    \caption{\textbf{RTA: reward-free control in two stages.} \emph{Stage 1} trains a VLM \emph{progress scorer} on shuffled clips with a GRPO objective that maximizes \emph{progress–time} Spearman $\rho$. \emph{Stage 2} uses the frozen scorer online: every $N$ steps, we score a window (with $L$ shuffles) and use the average $\rho\in[-1,1]$ as the \emph{only} reward for PPO-style learning.}
    \label{fig:stage1}
    \vspace{-10pt}
\end{figure*}

\vspace{-7pt}
\subsection{Stage 1: Listwise Progress Scorer via GRPO}
\label{sec:stage1}

\textbf{Expert clip batching.}
Given an expert video demonstration $\tau = (s_1, \dots, s_T)$, we segment it into frames $(s_1, \dots, s_{\tilde{K}})$, which we then use to construct training instances for progress prediction. Following~\citep{Ma2024GVL}, to avoid the shortcut “later is always better,” we \emph{anchor} the first frame and randomly shuffle the remaining frames:
\[
\begin{aligned}
&(s_{\mathrm{anc}},\, s_{\tilde{2}}, \ldots, s_{\tilde{K}}), \\
&(\tilde{2}, \ldots, \tilde{K}) = \texttt{permute}(2{:}K), \\
& s_{\mathrm{anc}} := s_1.
\end{aligned}
\]

Only the shuffled, non-anchor frames are scored; the anchor provides a fixed starting point for each training sequence. Fully shuffling frames can make the direction of progress ambiguous or overemphasize local appearance changes. Anchoring breaks this symmetry and encourages the model to evaluate true \emph{progress} with respect to the task, rather than merely tracking local appearance drift. The same anchoring and shuffling procedure is applied at inference time.

\textbf{VLM input and output.}
A vision–language model $f_\phi$ receives a sequence of $K$ frames as input and is prompted to generate, for each (shuffled) frame excluding the anchor, a reasoning trace that specifies its predicted \emph{progress rank} within the original sequence (higher = later). The canonical output format is:

\[
\begin{aligned}
&\texttt{Frame } i \texttt{:} \\
&\texttt{Frame Description: \ldots} \\
&\texttt{Rank: } p_i.
\end{aligned}
\]

Collecting these $K{-}1$ scalars yields the vector $\mathbf{p} = (p_2, \dots, p_K)$, where each $p_i$ denotes the predicted progress rank for frame $i$ (anchor fixed at position $1$ and never scored). All outputs are parsed using a strict regular expression. If parsing fails-for example, when the number of parsed scores does not equal $K{-}1$-we assign the minimum possible reward, $R_{\min} = -1$. For non-numeric or unparsable outputs, we treat parsing as failed. In case of ties equal ranks are permuted randomly.


\textbf{Listwise reward and target.}
Let $\mathbf{p}$ denote the predicted \emph{progress ranks} for the shown (shuffled) frames, and let $\mathbf{q}$ be their ground-truth temporal indices from original timestamps. We define the reward as
\begin{equation}
R \;=\; \spr\!\big(\mathbf{p},\,\mathbf{q}\big),
\label{eq:stage1_reward}
\end{equation}
where $\spr(\cdot,\cdot)$ is Spearman's rank correlation. Using Spearman on continuous progress ranks makes the objective insensitive to the specific scale of $p_i$ while rewarding correct monotone order.


\textbf{GRPO objective and optimization.}
We treat the VLM’s text generation process as a sequential policy over output tokens, and optimize a Group Relative Policy Optimization (GRPO) objective~\citep{GRPO2024}. The GRPO loss is:
\[
\frac{1}{|a|} \sum_{i=1}^{|a|} \min\Bigl( r_{i} \hat{A},\, \mathrm{clip}(r_{i}, 1-\epsilon, 1+\epsilon) \hat{A} \Bigr)
\]
where $a$ denotes the generated action (token) sequence, $r_{i} = \pi_\theta(a^i \mid s, a^{<i}) / \pi_{\theta_{\text{old}}}(a^i \mid s, a^{<i})$ is the importance weight, $\epsilon$ is a clipping parameter, and $\hat{A}$ is the relative group advantage. This advantage $\hat{A}$ is defined as in~\citet{GRPO2024}, based on the maximized expected reward:
\[
\max_{\phi} \;\mathbb{E}_{(\mathrm{segment},\,\mathrm{shuffle})} \;\mathbb{E}_{\mathrm{output} \sim f_\phi} \left[\, R \,\right]
\]
where $R$ is the progress–time Spearman correlation reward defined above. After convergence, the progress scorer $f_\phi$ is frozen for use in Stage~2.

\subsection{Stage 2: Online control from progress–time consistency}
\label{sec:stage2}

\textbf{Inference window, anchor, and reward.}
At each environment step $t$, construct the window $\mathcal{W}_{t} = (s_{t-m+1}, \dots, s_{t})$ with $m=\min(N,t)$, and designate the oldest state $s_{t-m+1}$ as the \emph{reference anchor} $s_{\mathrm{anc}}$. We query the frozen scorer only on \emph{query steps} ($t \bmod N = 0$ or $t=\mathcal{T}$); otherwise we set $r_t=0$. {\color{black}$N$ is a hyperparameter, which introduces a tradeoff between reward frequency and computational efficiency, since VLM inference makes per-step reward computation prohibitive.} On each query step, to reduce VLM cost and variance, draw $L$ independent permutations of the $m-1$ non-anchor frames (keeping $s_{\mathrm{anc}}$ fixed), obtain progress ranks from the scorer for each permuted window, compute their progress–time correlations, and average these correlations to yield the scalar reward $r_t$. Unless stated otherwise, we use $L=2$ and $N=15$.

\textbf{Policy optimization and normalization.}
We train the policy $\pi_\theta$ using {\color{black}policy gradient methods}, using $r_t$ as the sole reward signal (with no additional signals from the environment). The reward is bounded in $[-1,1]$ by construction. Since rewards are sparse (computed every $N$ steps) and bounded, we follow VL-DAC\citep{bredis2025enhancing} use Generalized Advantage Estimation (GAE) based on the observed $r_t$ sequence {\color{black}for VLM-based agents.}  

{\color{black} We use Qwen2.5-VL-7B as the backbone VLM for VL-DAC in all discrete environments. We choose a VLM backbone for game tasks because it can naturally handle combinations of actions described in text. This is important for the Kirby environment, which requires pressing multiple buttons at the same time, following the protocol of VideoGameBench~\citep{zhang2025videogamebenchvisionlanguagemodelscomplete}.}

{\color{black}To ensure that our results are driven by the reward design rather than biases of the VLM, we also evaluate a simple MLP backbone on Catrap games in combination with PPO~\cite{Schulman2017PPO}. These results are reported in Appendix~\ref{appendix:ablations}. For experiments with continuous control we use DrQv2~\citep{yarats2021masteringvisualcontinuouscontrol} as a policy.} For experiments with LOOP~\citep{loop}, see Appendix~\ref{appendix:loop}.


\section{Experiments}
\label{sec:experiments}

We evaluate RTA as a reward-free approach to vision–language policy learning in game-like and continuous environments. Agents receive only expert video demonstrations; no extrinsic rewards, environment APIs, or task annotations are used. 

The evaluation mirrors our two-stage method: (i) train and analyze a listwise VLM progress scorer that assigns each frame a progress rank; (ii) optimize a control policy using only the scorer’s progress–time correlation as reward. This setup tests two hypotheses: VLMs can reliably infer ordinal progress from video, and these signals alone suffice to train agents to solve complex tasks. 

{\color{black}Further details on the experimental setup, including data sources, task definitions, observation and action processing, and training and compute details, are provided in Appendix~\ref{appendix:setup}.}

\subsection{Stage~1: Progress-Scorer Learning Dynamics and Generalization}
\label{sec:exp:stage1}

\textbf{Learning dynamics.} We begin by verifying that a single VLM-based scorer can learn to score progress within shuffled expert clips across different levels, and we investigate its ability to generalize across levels, games, and domains. We run Stage~1 on both individual Catrap levels and for complete games playthroughs. For almost all videos algorithm was able to converge to Spearman $\rho > 0.9$ in at most 300 steps. For more details refer to the appendix~\ref{appendix:stage1_learning_dyn}.

\begin{wrapfigure}{r}{0.45\textwidth}
  \vspace{-12pt}
  \centering
  \includegraphics[width=\linewidth]{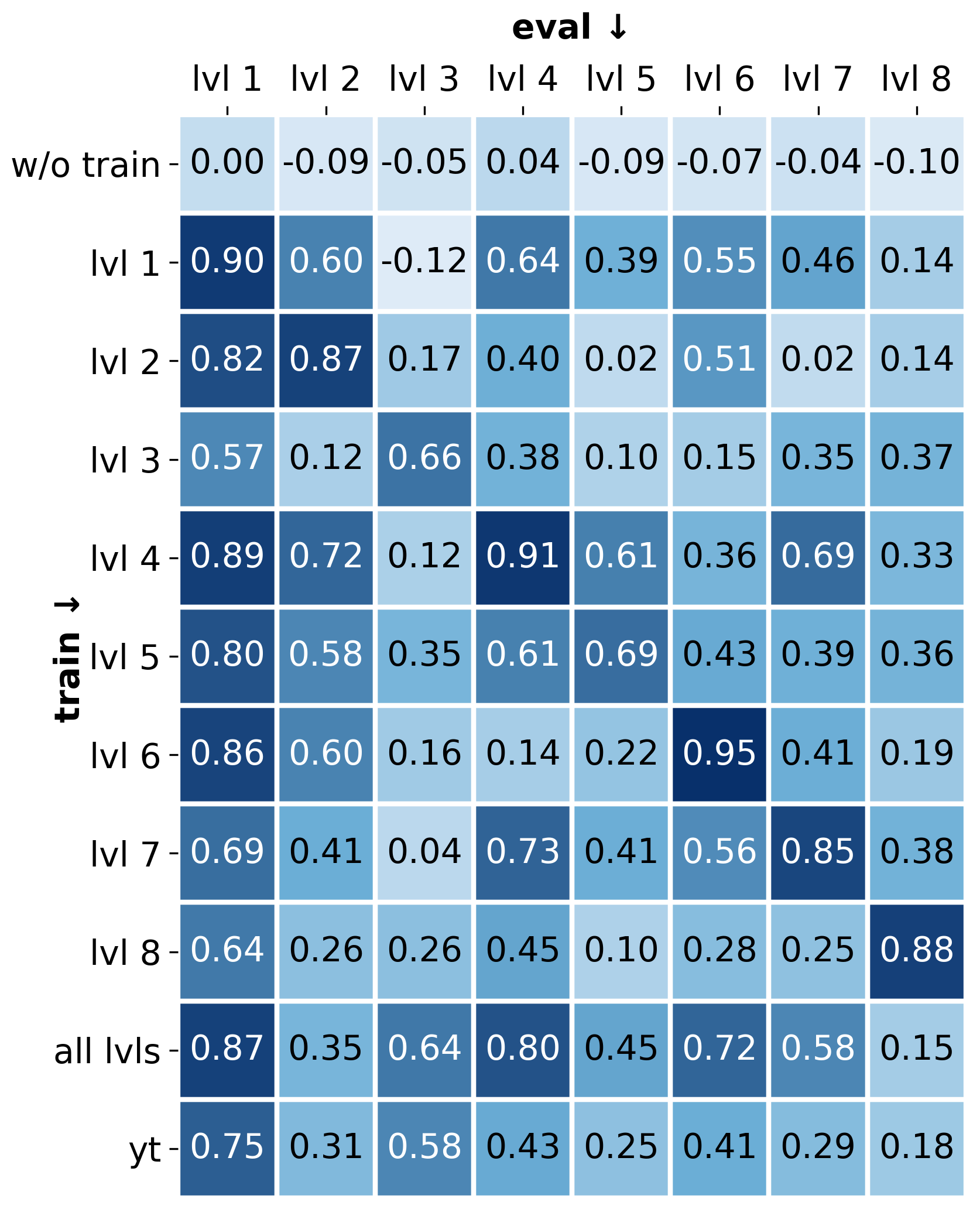}
  \caption{\textbf{Cross-level transfer of ordinal progress.} Heatmap of validation \emph{progress–time} $\rho$ for source (rows) $\rightarrow$ target (columns). Diagonals are in-level; off-diagonals show transfer. Pooled training (all levels or diverse games) transfers best; simpler Level 1 generalizes from most sources. {\color{black}VLM-based scorer trained on a single level (or YouTube dataset) recovers progress from shuffled sequences across different levels, showing higher correlation than an untrained model.}}
  \label{fig:xlevel_heatmap}
  \vspace{-3em}
\end{wrapfigure}

\textbf{Cross-level generalization.}We train the progress scorer on a \emph{source} set-either a single level $i$, a full-game playthrough, or a pool of 70 scraped GameBoy playthroughs-and evaluate on a disjoint \emph{target} level $j$. Performance is the Spearman correlation between predicted progress and time (\emph{progress–time} $\rho$). Figure~\ref{fig:xlevel_heatmap} shows the mean validation $\rho$ for all source–target pairs $(i \to j)$: diagonal cells (within-level) give upper bounds, while off-diagonals measure transfer.

Both per-level and pooled setups achieve high in-distribution $\rho$, however transfer is notably asymmetric: scorers trained on any level generalize especially well to level~1, likely due to its lower complexity. The strongest overall transfer comes from pooled training on all levels, indicating that a single, shared scorer can support domain-general inference-from full-game playthroughs to individual levels. Training on a diverse mix of 70 GameBoy playthroughs also yields solid zero-shot performance on held-out levels, despite limited direct exposure.


\vspace{-7pt}

\subsection{Stage 2:~Control from progress–time consistency}
\label{sec:exp:stage2}

We now deploy the frozen progress scorer in the environment and train a policy using policy-gradient methods, optimizing \emph{only} the progress–time correlation reward $r_{t}$ (see Sec.~\ref{sec:stage2}), with no access to any extrinsic task rewards. 






\textbf{GameBoy results.}
We conduct experiments on Catrap level 2, level 4, level 6, and the Kirby game. Levels 4 and 6 require advanced reasoning and backtracking from the model, as solving these levels involves overcoming dead ends. Kirby is a long-horizon control game. In the absence of explicit environment rewards, there are several approaches for current VLM and LLM-based control is to train  Rank2Reward (which has not previously been applied to VLMs) and VLM-RM. Additionally, we compare our method against an untrained stage-1 model, which is equivalent to the GVL~\citep{Ma2024GVL} formulation, {\color{black} and with Gemini-3.1-Thinking~\cite{gemini2025} as a backbone}.  A comparison of success rates can be found in Table~\ref{table:baseline_comparison}, where RTA outperforms the other methods. 

Additionally, to show that RTA provides a stable reward signal for agent learning, we report the Pearson correlation between mean cumulative reward and pass@5 after each training update for the best-performing run (in terms of success rate) for every method. Results are presented in Table~\ref{table:reward_correlation}. Appendix~\ref{appendix:lvl2_curve} shows the evolution of the per-query \emph{progress–time} Spearman reward during training on Catrap Level 2. We plot both the progress–time reward and the success rate for direct comparison; both metrics increase steadily and in pair, further validating the coherence and shaping power of our reward signal. 


{ 
\begin{table*}[t]

\caption{Comparison of success rates from different approaches. Means and std are computed over 5 seeds. Oracle reward is binary reward according to success of the end of level. {\color{black}RTA achieves the strongest performance across all levels except Kirby. Notably, it is the only method that attains a non-zero success rate on Kirby, despite not using any environment-provided rewards.} }

\centering\centering
\begin{tabular}{lcccc}
    \toprule
    Task       & Level 2  & Level 4 & Level 6 & Kirby (level 0)\\
    \midrule
    GVL  &  0.47 ± 0.25 & 0.00 ± 0.00 & 0.04 ± 0.08 & 0.00 ± 0.00 \\
    {\color{black}GVL-Gemini} & 0.27 ± 0.09 &	0.00 ± 0.00 & 0.00 ± 0.00 & 0.00 ± 0.00\\ 
    VLM-RM  &  0.40 ± 0.28 &	0.16 ± 0.20 &	0.08 ± 0.16 &	0.00 ± 0.00\\
    VLM-RM$_{reg}$($\alpha=0.5$) & 0.44 ± 0.32 &	0.00 ± 0.00	&0.08 ± 0.16	&0.00 ± 0.00 \\
    Rank2Reward  &  0.60 ± 0.28&	0.20 ± 0.00&	0.13 ± 0.09&	- \\
    Oracle reward & 0.50 ± 0.22&	0.07 ± 0.09	&0.20 ± 0.00&	\textbf{0.40 ± 0.28} \\
    \midrule
    
    RTA (w stage1 training)        &   \textbf{1.00 ± 0.00} & \textbf{0.72 ± 0.35} & \textbf{0.32 ± 0.27} & 0.07 ± 0.09   \\
    \bottomrule
\end{tabular}

\label{table:baseline_comparison}
\vspace{-3pt}
\end{table*}
}

\begin{table*}[h]
\centering
\caption{{  Cross-domain evaluation of RTA using different video sources for Stage-1 training. Results in this table report mean ± std across 3 seeds.} {\color{black} Scorers trained on different visual domains still provide meaningful rewards, enabling agents to complete levels as well as or better than the baselines reported in the Table~\ref{table:baseline_comparison}.}}
\begin{tabular}{lcccc}
    \toprule
    Source        & Level 2  & Level 4 & Level 6 & Kirby (level 0) \\
    \midrule
    Youtube       & 1.00 ± 0.00 & 0.47 ± 0.25 & 0.60 ± 0.28 & 0.20 ± 0.16 \\
    Full Catrap   & 1.00 ± 0.00 & 0.47 ± 0.38 & 0.60 ± 0.28 & 0.07 ± 0.09 \\
    Full Kirby    & --          & --          & --          & 0.07 ± 0.09 \\
    {\color{black}MetaWorld} & 1.00 ± 0.00 & 0.87 ± 0.19 & 0.53 ± 0.19 & 0.13 ± 0.09 \\ 
    {\color{black}Coin (AssembleSofa)} & 1.00 ± 0.00 & 0.07 ± 0.09 & 0.20 ± 0.28 & 0.00 ± 0.00 \\
    \bottomrule
\end{tabular}

\label{table:cross_domain}
\vspace{-3pt}
\end{table*}




\textbf{Continuous control.} To evaluate RTA beyond discrete tasks, we test it on PointMaze-UMaze, and selected MetaWorld environments, which present more challenging long-horizon dynamics. For these tasks, we adopt DrQv2~\cite{yarats2021masteringvisualcontinuouscontrol} as the backbone and compare RTA with Rank2Reward (R2R). Both methods were trained for one million environment steps. Additionally, we ablate the effect of combining a GAIL-style reward with the ranking-based reward, following a similar approach to R2R, where the GAIL reward is added to the ranking-only reward to improve overall learning quality.

\begin{figure}[t]
    \centering
    \includegraphics[width=0.7\linewidth]{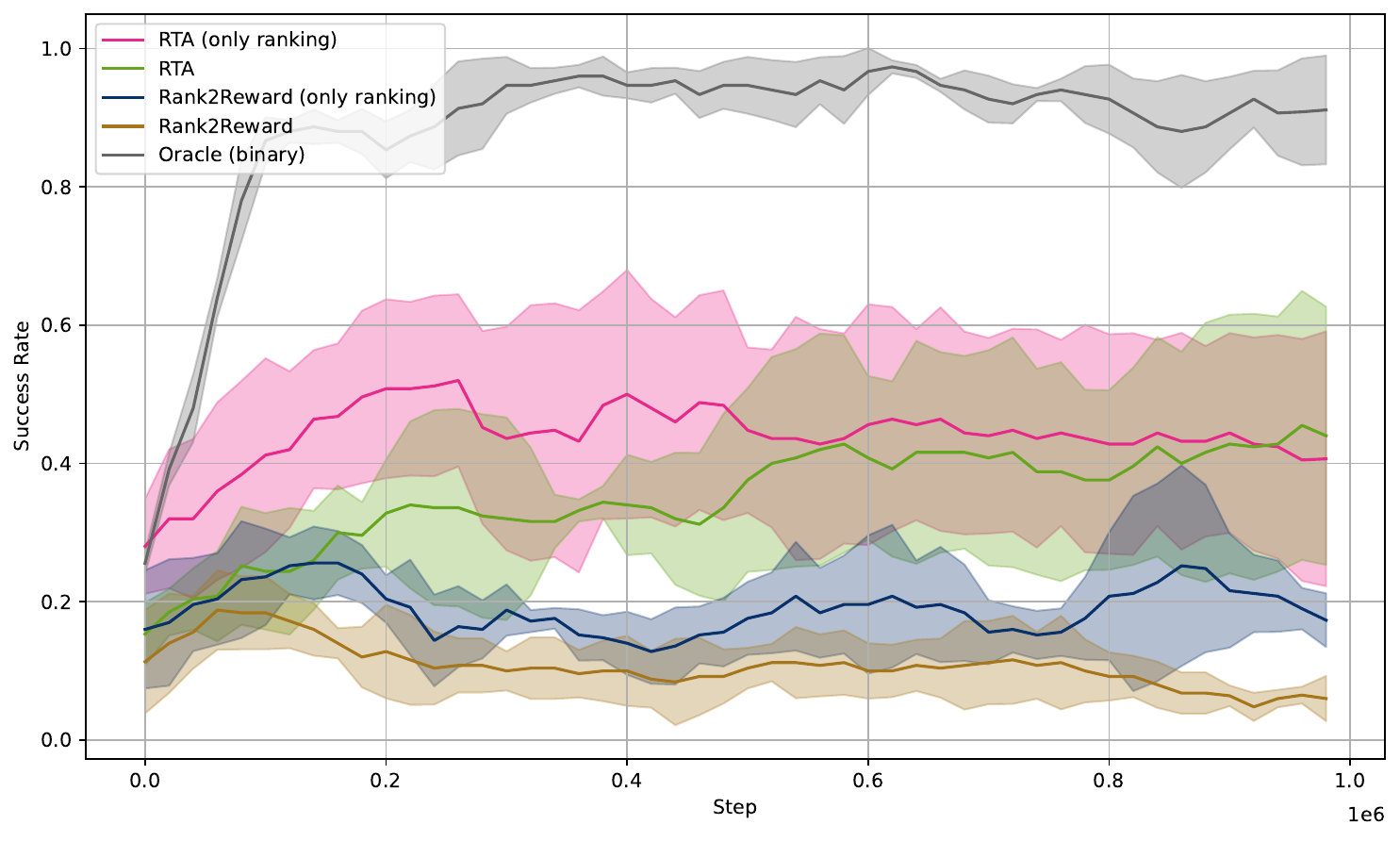}
    \caption{\textbf{PointMaze-UMaze results}. Results in this table report mean ± std across 5 seeds, smoothed with a moving average (window size 5) for visual clarity. {\color{black}RTA outperforms R2R both with and without GAIL. The oracle reward is included for comparison, although it is not directly comparable to our method.}}
    \label{fig:pointmaze}
    \vspace{-5pt}
\end{figure}

As shown in Figure~\ref{fig:pointmaze}, RTA consistently outperforms R2R on UMaze, both in the ranking-only and mixed reward settings, in terms of success rate. On MetaWorld tasks, RTA also surpasses R2R under the ranking-only reward, though it still underperforms methods that incorporate an additional classifier-based (GAIL-style) reward. This behavior likely arises because MetaWorld environments require extensive exploration, and RTA occasionally assigns high VOC-scores to trajectories with plausible dynamics that are nevertheless unsuccessful. 

\begin{figure*}[t]
    \centering
    \includegraphics[width=0.8\linewidth]{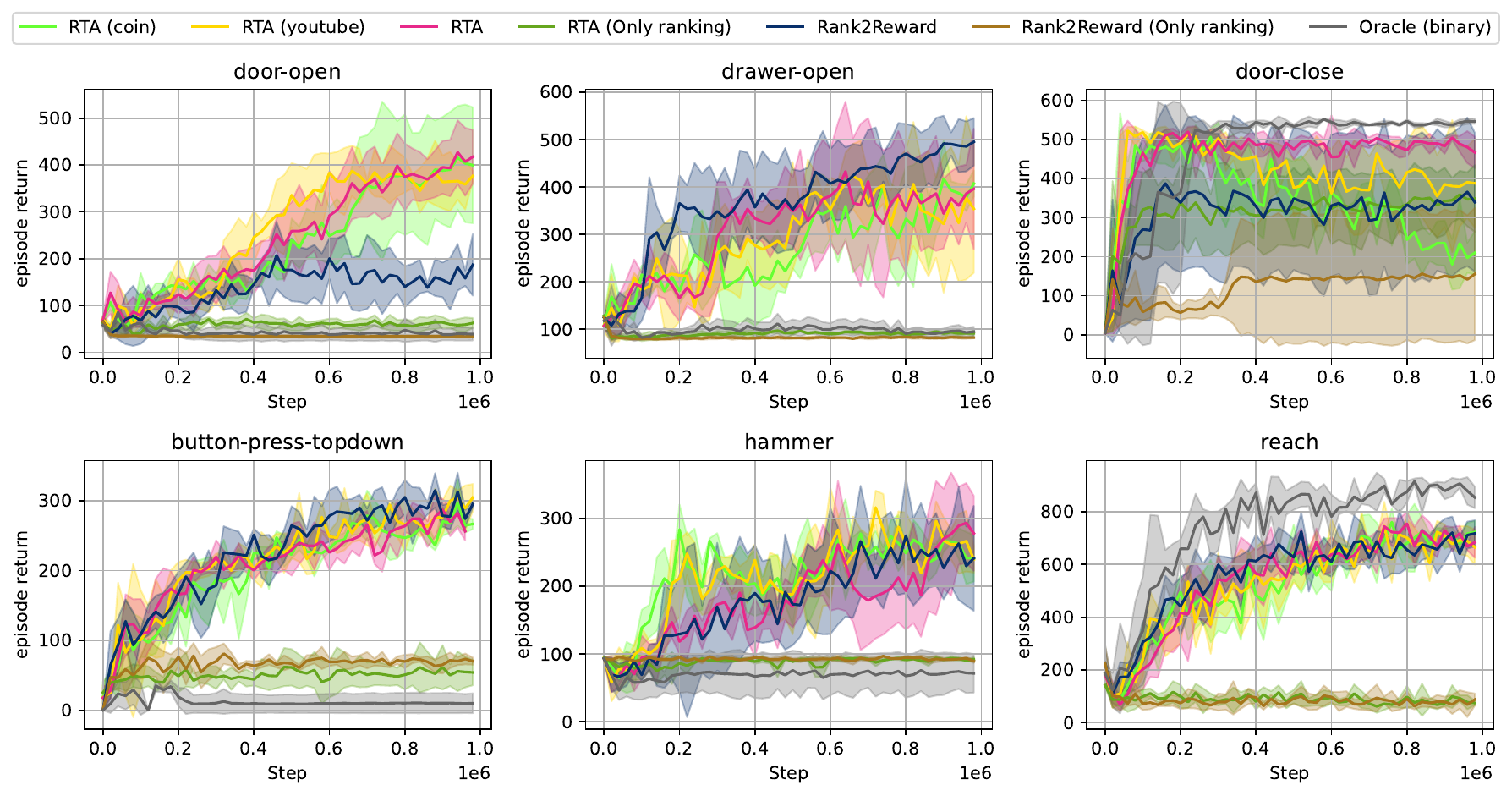}
    \caption{{\color{black} \textbf{MetaWorld different tasks results}. Stage-1 RTA is trained once on a mixed dataset of demonstrations, while R2R is trained separately per task. We additionally evaluate Stage-1 scorers trained on different data sources, including COIN and YouTube datasets. Results report mean ± std over 4 seeds. RTA + GAIL achieves comparable returns to R2R across all tasks, and outperforms it on \texttt{door-open} and \texttt{door-close}. Unlike R2R, RTA requires no per-task fine-tuning, and the Stage-1 scorer generalizes well across data sources and unseen environments, enabling successful task completion.}}
    \label{fig:metaworld}
    \vspace{-1.5em}
\end{figure*}

However, R2R relies on pretraining a separate model for each task, whereas RTA allows a single pretrained model to be reused across multiple tasks, offering greater scalability. Success rates and comparisons to original dense rewards for MetaWorld tasks are provided in Figure~\ref{fig:metaworld}. 

{\color{black} We also analyze the efficiency of RTA compared to R2R under a matched compute budget (see Appendix~\ref{app:efficiency}). The results show that RTA remains competitive across all tasks and outperforms R2R on the majority of them, indicating that its advantage is not due to increased computational resources.}


\textbf{Cross-domain generalization.} \label{para:cd}To examine the generalization of the Stage~1 scorer across video sources, we carried out additional experiments in which RTA is trained on (a) a single full-game Catrap playthrough, (b) a mixture of YouTube PyBoy playthroughs from multiple games and players (excluding Catrap and Kirby), and (c) a Kirby playthrough. These sources differ in visual appearance (compression artifacts, overlays, distinct play styles, etc.) but share the same underlying game dynamics. Results are shown in Table~\ref{table:cross_domain}. 

{\color{black}To further show the generalization ability of RTA, we run additional experiments where the Stage 1 ranker is trained on data with very different visual dynamics. In particular, we train the ranker on expert trajectories from MetaWorld and on AssembleSofa videos from the COIN dataset.~\cite{tang2019coin}. } 

{\color{black}We also show that the generalization ability is preserved in MetaWorld environments when the ranker is trained on different source videos, including a YouTube GameBoy playthrough collection and AssembleSofa videos from the COIN dataset. Fig.~\ref{fig:metaworld}).}

\textbf{Ablations.} We conduct several ablations on window size, reward frequency, and reward averaging. In addition, we experiment with using an MLP backbone for the agent in Stage~2 instead of a VLM, demonstrating that the choice of policy model does not affect training outcomes and that performance gains are primarily due to reward design. Finally, we compare giving rewards after every $N$ steps versus providing a single reward at the end of the trajectory (with samples taken uniformly over the trajectory). Further details are provided in Appendix~\ref{appendix:ablations}.

\begin{figure*}[t]
    \centering
    \includegraphics[width=\linewidth]{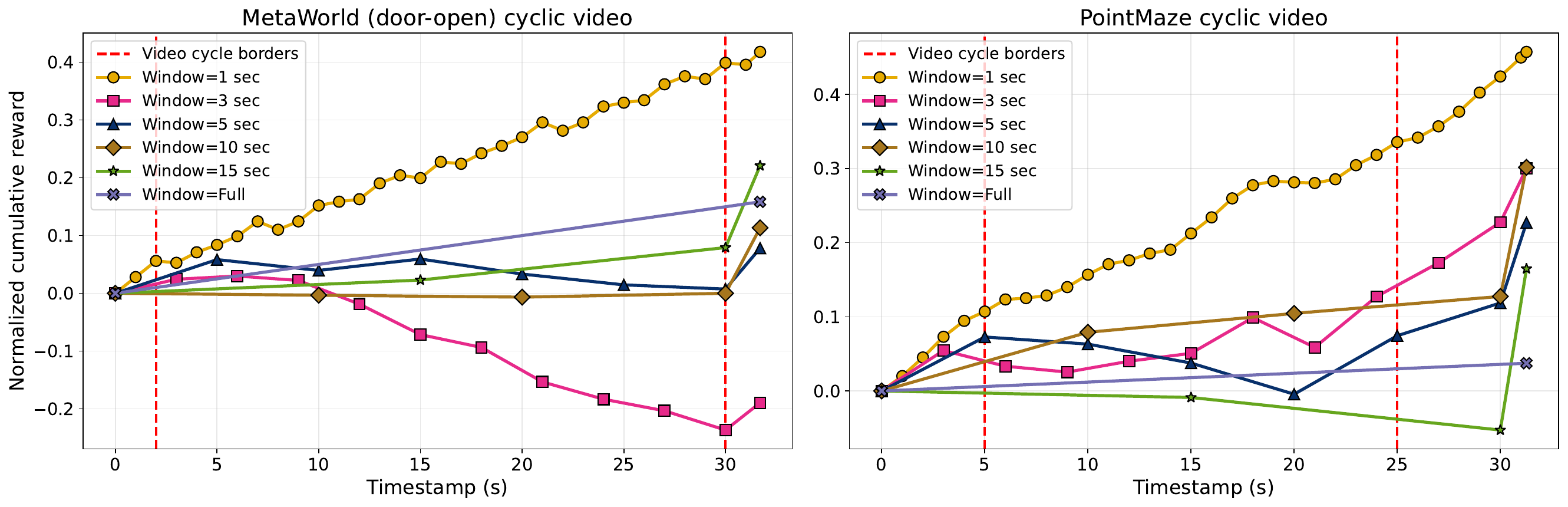}
    \caption{{\color{black} \textbf{Effect of window size on the reward signal on cyclic trajectory.} Small windows produce increasing reward despite no true progress, reflecting local consistency. Medium windows better capture the absence of global progress, while large windows yield near-zero signal due to averaging. All window sizes assign high reward to the final segment where the goal is achieved.}}
    \label{fig:loop}
    \vspace{-1em}
\end{figure*}

{\color{black}\textbf{Robustness to temporal structure and window size.}
We study how the windowed rank-correlation reward behaves under different temporal scales using cyclic trajectories composed of forward and reversed segments (Appendix~\ref{app:cyclic}). This construction explicitly removes global monotonic progress, allowing us to isolate the inductive bias of the reward signal.

In Fig.~\ref{fig:loop}, we report cumulative normalized reward over time using non-overlapping windows of varying length. We find a clear scale-dependent behavior: (i) very small windows are overly sensitive to local temporal consistency and can produce monotonically increasing reward even in cyclic trajectories; (ii) intermediate windows suppress these local artifacts and better reflect the absence of global progress; and (iii) large windows lead to signal dilution due to temporal averaging.

Despite these differences, all window sizes consistently assign high reward to the final segment where the goal is achieved, indicating that the signal preserves sensitivity to true task completion. Overall, this reveals a stable intermediate regime in which windowed rank-correlation acts as a reliable progress estimator, while also exposing its limitations under extreme temporal scales.}

\section{Discussion}
\label{sec:discussion}
\vspace{-7pt}
By learning policies directly from expert videos using a progress–time correlation signal, our work suggests a scalable approach to reward-free learning. This reduces the need for manually designed reward functions and may help avoid reward hacking and unintended behaviors. 

We leave detailed parsing issues to prior work. Stage 1 ranking is similar to reasoning-style VLM supervision, where parsing and ranking stability have been studied~\citep{shao2024deepseekmathpushinglimitsmathematical}. Similar issues in Stage 2 VLM-based learning have also been discussed in multi-turn settings~\citep{zhai2024finetuninglargevisionlanguagemodels}.

We also show examples of correct and incorrect scoring, including cases where an undertrained Stage 1 model leads to wrong reward assignment (see Appendix~\ref{appendix:examples}). 

Our method supports scoring long trajectories by sampling frames and leveraging VLM reasoning to infer temporal order, which allows application beyond strictly monotonic tasks.

\textbf{Limitations.}
The approach relies on sufficiently diverse expert video data; limited coverage or dataset bias may lead to level-specific overfitting and weaker cross-level transfer, as reflected in asymmetric Stage~1 generalization.

While RTA does not assume globally monotonic progress, the windowed rank-correlation signal exhibits scale-dependent behavior: small windows can overemphasize local temporal consistency, while large windows may dilute meaningful structure. Our cyclic analysis (Appendix~\ref{app:cyclic}) and ablations (Appendix~\ref{appendix:ablations}) show that an intermediate window regime is most stable across environments. Future work could further improve robustness in highly non-monotonic settings via multi-scale or hierarchical temporal aggregation, or explicit subgoal-aware reward modeling.

Combining RTA with complementary signals (e.g., a GAIL-style reward) may further mitigate cases where the scorer assigns high rewards to out-of-domain or non-progressing trajectories.

Current VLM backbones also introduce latency despite caching and striding, motivating more efficient or distilled models. 

Overall, ordinal video-based rewards remain a promising direction for scalable generalist control.

\vspace{-7pt}

\section{Conclusion}
\vspace{-7pt}

We present Rank-Then-Act (RTA), a simple two-stage method that teaches vision-language models (VLMs) to act using only expert video and progress ranks (via Spearman correlation), with no extrinsic rewards. In both challenging discrete-action games and long-horizon continuous environments, this purely reward-free signal trains agents that achieve high success rates and outperform strong baselines. 

Because real-world rewards are often sparse or hard to design, learning from expert video alone offers a scalable path to multimodal agency while reducing risks like reward hacking. 

We see this as a step toward robust, generalist agents that can understand and act in open, dynamic, and underspecified environments.

\bibliography{example_paper}
\bibliographystyle{plainnat}

\newpage
\appendix
\onecolumn

\section{Appendix}

\subsection{Experiments with LOOP}
\label{appendix:loop}
For experiments with LOOP-an extension of GRPO to multi-turn settings-we set the episode length to $\mathcal{T}=N$ and compute the reward only at the end of each episode, assigning the same advantage to all output tokens. Because this limits long-horizon environment interaction, we restart the trajectory from a new starting point whenever the average episode reward $r_{\mathcal{T}}$ across sampled traces exceeds a threshold $\tau$. At each reset, we select the trace with the highest reward and repeat the process.

\begin{figure}[t]
  \centering
  \includegraphics[width=0.8\linewidth]{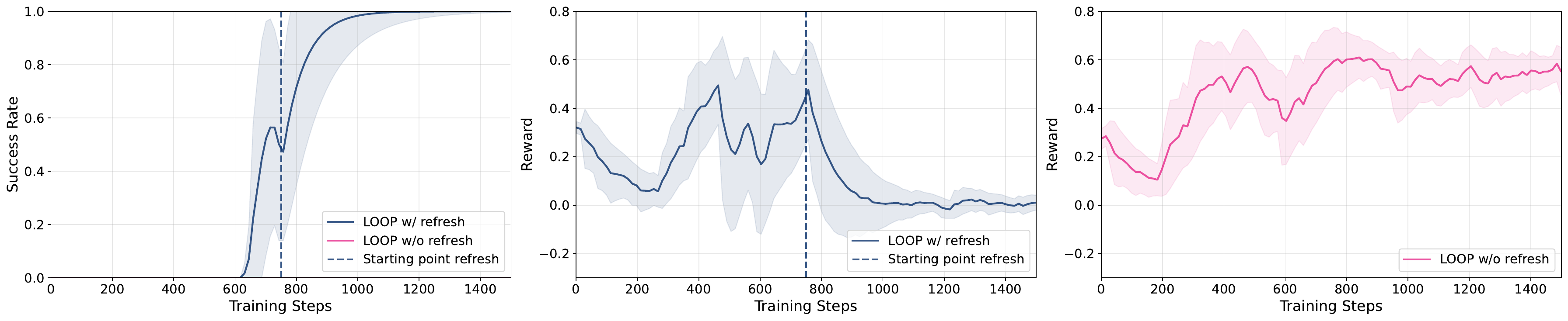}
  \caption{\textbf{LOOP benefits from starting-point refreshing.} Without refreshing, rewards improve but may not reach success when $N$ is shorter than the task. Refreshing from the best terminal state after threshold $\tau$ enables completion; brief post-refresh dips arise if $N$ exceeds remaining steps.}
  \label{fig:stage2_loop}
\end{figure}

Figure~\ref{fig:stage2_loop} presents the per-query \emph{progress–time} Spearman reward during training with LOOP~\citep{loop}. We report two reward curves: one for training with \textbf{starting point refreshing} enabled, and another without refreshing.

When the model is trained without refreshing the starting point, rewards increase steadily, but there is no corresponding rise in success rate. This occurs because the default window length $N$ is insufficient for the agent to complete the full task within a single episode. To address this, we introduce \textbf{starting point refreshing}: after achieving a threshold $\tau$, we select the trajectory with the highest reward and begin the next rollout from the final state of that trajectory. This approach enables the algorithm to achieve high success rates, as reflected by the success curve.

However, if the new rollout’s length $N$ exceeds the number of steps required to finish the level, this can occasionally produce a temporary decline in reward after refreshing, as seen in the post-refresh segments of the plot. This effect is due to the agent finishing the level quickly, leaving extra rollout steps with little meaningful scoring signal.

\subsection{Stage 1: Progress-Scorer Learning Dynamics }
\label{appendix:stage1_learning_dyn}

\begin{figure}[t]
  \centering
  \includegraphics[width=\linewidth]{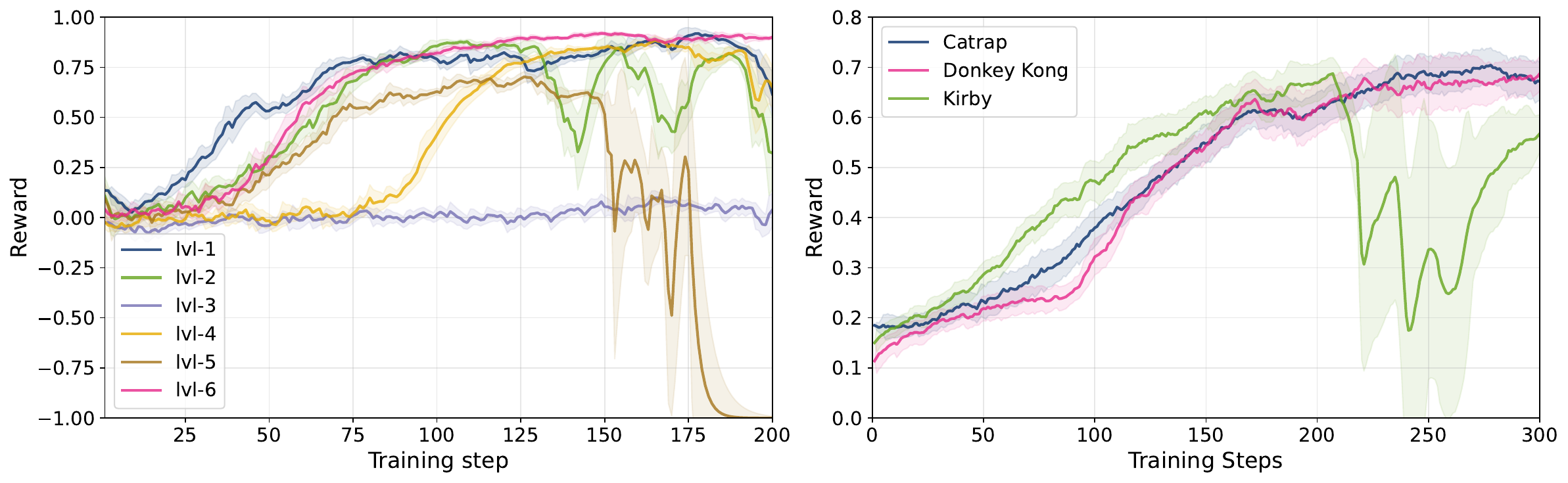}
  \caption{\textbf{Stage 1 learns reliable ordinal progress.} \textbf{Left:} per-level Catrap training shows fast rises in \emph{progress–time} $\rho$ with slower convergence on intricate levels. \textbf{Right:} training on full playthroughs across games also converges, with slightly lower plateaus due to non-informative segments (e.g., menus).}
  \label{fig:stage1_by_level_and_playthroughs}
\end{figure}

\textbf{Per-level training curves.}
We first train the scorer on individual Catrap levels. The left panel of Fig.~\ref{fig:stage1_by_level_and_playthroughs} shows the GRPO training curves for the first six levels. Most levels exhibit rapid convergence within 200 steps, achieving asymptotic \emph{progress–time} Spearman $\rho$ near $0.9$. Some training runs diverge after reaching high performance-a phenomenon also observed in extended GRPO training, and consistent with known GRPO instabilities. All training videos in this experiment were extracted under human supervision.

 As shown in ~\ref{appendix:stage1_level3}, level 3 displays alternating periods of convergence and divergence, suggesting that even in levels with less clear task goals, progress can be learned, albeit with greater sensitivity to data preparation and training duration.

\textbf{Full-playthrough training.}
Next, we pool all Catrap levels (1–8) and train a \emph{single} scorer across this part of game. The right panel of Fig.~\ref{fig:stage1_by_level_and_playthroughs} shows training $\rho$ on a mixed-level and games held-out set. In these experiments, we train the model for 300 steps on three different games to achieve and sustain high asymptotic quality.

Pooling all levels increases data diversity and produces smoother training dynamics, though the final performance plateau is slightly lower than in most single-level runs. This drop is likely due to segments with static or non-informative frames (such as menus) present in full playthroughs. Importantly, this experiment demonstrates that a single progress scorer can be trained end-to-end from full-game videos, obviating the need for level-specific tuning and opening a path toward scalable, domain-general learning. Additional results confirm that per-game training also provides robust performance on level-specific validation.

\subsection{Setup}
\label{appendix:setup}

\textbf{Expert videos.}
For Stage~1, we use human \emph{Catrap} playthroughs (levels 1–8; successful episodes), one \emph{full-game} Catrap run, and 70 additional GameBoy playthroughs scraped from YouTube for discrete environments and 100 trajectories collected from expert agents for MetaWorld~\cite{mclean2025metaworld} and PointMaze~\cite{gymnasium_robotics2023github} tasks. Videos are split into fixed-length clips of $K$ consecutive frames and converted to the \emph{anchor+shuffle} interface (Sec.~\ref{sec:stage1}). To standardize inputs and avoid cues such as completion time or lives, GameBoy frames are cropped from $640{\times}360$ to $640{\times}330$.  

\textbf{Tasks and levels.} For control evaluation, we consider multiple Catrap levels (L2, L4, L6, with progressive difficulty) and a second game, Kirby (part until the first boss is met, corresponding to the first checkpoint in VideoGameBench~\citep{zhang2025videogamebenchvisionlanguagemodelscomplete}, we'll call it level-0 for simplicity), in the PyBoy emulator. {\color{black}Catrap is a logically complex puzzle requiring spatial planning. Kirby is a standard VLM benchmark from VideoGameBench. Both use the simplest available levels that require at least 15 steps for successful completion.} For MetaWolrd we choose tasks \texttt{door-ope}, \texttt{drawer-open}, \texttt{door-close},  \texttt{button-press-topdown}, \texttt{hammer}, \texttt{reach}, and for PointMaze we test our setup on UMaze task. Each task is defined by a goal description and termination condition, and we train separate policies per task while reusing a trained (from per-task to general) Stage-1 progress scorer. This setup stresses both cross-level and cross-game reuse of the scorer, and enables us to probe robustness of the progress–time correlation reward beyond a single level.

\textbf{Observations and actions.}
For Stage~2 policy learning in PyBoy games, we render environment observations at $160 \times 144$ pixels; for policy inputs, frames are centrally cropped and resized to $160 \times 120$. The action space consists of the four D-pad directions (\texttt{up}, \texttt{down}, \texttt{left}, \texttt{right}) defined by PyBoy's discrete control interface. To ensure consistency with Stage~1, we resize frames back to the original training resolution before querying the frozen scorer $f_\phi$ to obtain progress ranks, which we then correlate (\emph{progress–time} Spearman) with timestamps. For Kirby we allow pressing multiple buttons simultaneously, following the VideoGameBench Light setup. For continuous environments we follow the setup from Rank2Reward~\cite{Yang2024Rank2Reward}.


\textbf{Technical details.}
For Stage1 and Stage 2 experiments with VL-DAC as a backbone policy, we train only the LoRA adapters (rank 16, alpha 32) on top of the provided models. For Stage~1 scorer training, each reported experiment requires 4 H100 GPUs, and 200 training steps take approximately 2 hours. For Stage~2 policy learning {\color{black}(with RTA as a reward model)}, 1 H100 GPU is required, and training takes up to 24 hours. Detailed hyperparameter settings are provided in Appendix~\ref{appendix:hypers}.

\subsection{Experimental details}
\label{appendix:hypers}

\begin{table*}[h]
    \caption{\textbf{Stage 2 (VL-DAC) hyperparameters.} Policy learns from the windowed \emph{progress–time} $\rho\in[-1,1]$ computed every $N$ steps (default $N\!=\!15$, $L\!=\!2$). $\gamma_g,\lambda_g$ are GAE parameters; LR follows a cosine decay.}
    \centering
    \begin{tabular}{l|c|c|c}
        \toprule
        Hyperparameter & Values  \\
        \midrule
        Env.steps & 57600    \\
        Learning Rate (init $\rightarrow$ final) & 1e-5 $\rightarrow$  5e-7    \\
        Scheduler & cosine   \\
        GAE $\lambda_g$ & 0.95 \\
        $\gamma_g$ & 0.99  \\
        Value Loss Coeff. & 0.15\\
        KL $\beta$ & 0.05  \\
        Policy Freeze (steps) & 2  \\
        Grad Accum. Steps & 32  \\
        Mini-batch Size & 1   \\
        PPO Epochs & 2  \\
        Obs. Image Length & 5  \\
        Rollout Size & 256 \\ 
        Max Episode Steps & 64 \\
        Temperature & 0.2 \\
        \bottomrule
    \end{tabular}
    
    
    \label{tab:hyperparams_vldac}
\end{table*}

\begin{table*}[ht]
    \caption{\textbf{Stage 2 (LOOP) hyperparameters.} Multi-step GRPO variant using the same correlation-only reward. Training optionally applies \emph{starting-point refreshing} when reward exceeds threshold $\tau$ to enable long-horizon completion.}
    \centering
    \begin{tabular}{l|c|c|c}
        \toprule
        Hyperparameter & Values \\
        \midrule
        Algorithm steps & 225    \\
        Learning Rate (init $\rightarrow$ final) & 1e-5 $\rightarrow$  1e-6   \\
        Scheduler & linear with warmup    \\
        Num. warmup steps & 10 \\
        KL $\beta$ & 0.05 \\
        Grad Accum. Steps & 32 \\
        Mini-batch Size & 1   \\
        PPO Epochs & 1 \\
        Obs. Image Length & 5  \\
        Rollout Size & 120 \\
        $\tau$ threshold & 0.5 \\
        $\tau$ threshold steps & 3 \\
        K & 4 \\
        Temperature & 1 \\
        \bottomrule
    \end{tabular}
    
    \label{tab:hyperparams_loop}
\end{table*}

\begin{table*}[ht]
    \caption{\textbf{Stage 1 scorer hyperparameters.} GRPO on anchor+shuffle clips; the VLM predicts per-frame \emph{progress ranks} and maximizes \emph{progress–time} Spearman $\rho$. After convergence, the scorer is frozen for Stage 2.}
    \centering
    \begin{tabular}{l|c|c|c}
        \toprule
        Hyperparameter & Values  \\
        \midrule
        Algo steps & 200-400    \\
        Learning Rate  & 1e-5    \\
        Scheduler & constant with warmup  \\
        Num. warmup steps & 10 \\
        Grad Accum. Steps & 16  \\
        Mini-batch Size & 1   \\
        GRPO Epochs & 1  \\
        Obs. Image Length & 15  \\
        K & 4 \\
        Temperature & 1 \\
        \bottomrule
    \end{tabular}
    
    \label{tab:hyperparams_stage1}
\end{table*}

\textbf{Hyperparameters (summary).}
\textbf{Stage~1:} GRPO on a tuned VLM; the first frame is anchored, and the remaining $K{-}1$ frames are shuffled; the listwise reward is $R=\spr(\mathbf{p},\mathbf{q})$, where $\mathbf{p}$ denotes \emph{progress ranks} and $\mathbf{q}$ denotes temporal indices (timestamps); a moving-average baseline is used.\
\textbf{Stage~2:} Reward frequency $N=15$; $L=2$ shuffled-window evaluations per query; window length $m$ as in Sec.~\ref{sec:stage2} (we ablate $N$ and $L$); per-episode standardization is used only for advantage estimation.\
The policy and value networks in Stage~2 are trained with VL-DAC~\citep{bredis2025enhancing} and a multi-step variant of GRPO, LOOP~\citep{loop}. Importantly, no environment rewards are used at any stage.
Additionaly we employ kl control on stage1, we set target kl to $0.1$.
Also we list hyperparameters for stage1 and stage2 in tables \ref{tab:hyperparams_loop} \ref{tab:hyperparams_vldac} \ref{tab:hyperparams_stage1}

\subsection{Ablations}
\label{appendix:ablations}

In this section, we present an expanded set of ablations evaluating both algorithmic hyperparameters of RTA and architectural choices for the downstream policy. These studies are designed to isolate the contribution of each component and determine the robustness of RTA across implementation variations. All experiments are conducted on Catrap environments and results are averaged over multiple seeds.

 \textbf{Policy backbone choice}. We further evaluate how RTA interacts with different policy backbones, comparing a VLM with a lightweight MLP trained with PPO. Table \ref{table:ppo_downstream} summarizes the results for two downstream settings: providing RTA reward every $N=15$ steps and providing reward only at the end of an episode.

We observe that the optimal schedule differs between the VLM and MLP policies, plausibly due to different exploration dynamics: an initialized MLP benefits from termination step RTA feedback (we reward full trajectory), whereas the VLM policy often benefits from shorter-window shaping. Overall, these results are consistent with the view that RTA’s gains are driven by the reward signal’s ordering stability rather than a particular policy architecture.

\begin{table*}[h]
\caption{Downstream PPO evaluation results. Mean ± std over 3 seeds for VLM-based agents and 5 seeds for MLP agents. {\color{black}RTA maintains a high success rate even with an MLP backbone, indicating that its performance stems from the reward design rather than biases in VLMs. }}
\centering
\begin{tabular}{lccc}
    \toprule
    Task                                   & Level 2        & Level 4        & Level 6    \\
    \midrule
    VLM + RTA (reward every 15 steps)       & 1.00 ± 0.00    & 0.60 ± 0.28   & 0.33 ± 0.34 \\
    VLM + RTA (only-end reward)            & 0.33 ± 0.19    & 0.00 ± 0.00   & 0.00 ± 0.00 \\
    \midrule
    MLP + RTA (reward every 15 steps)       & 0.79 ± 0.14    & 0.27 ± 0.21   & - \\
    MLP + RTA (only-end reward)            & 1.00 ± 0.00    & 1.00 ± 0.00  & 1.00 ± 0.00  \\
    \bottomrule
\end{tabular}

\label{table:ppo_downstream}
\end{table*}

\textbf{Window mechanics}. We vary the window length $m$ around our default. Across Catrap levels, we find that performance is stable with the default value $m=15$ and higher, while very short windows (too few frames) perform worse, as expected. For larger windows than default, convergence tends to be the same, and inference is slower due to the larger context processed by the reward model. We fix the reward frequency to 15 and $L$ to 2.

\textbf{Reward frequency}. We vary the frequency at which reward is provided to the learning process. We find that this parameter does not qualitatively affect stability, and its effect on convergence speed appears to depend more on optimization dynamics. We fix the window length to 15 and $L$ to 2.

\textbf{Number of shuffles L}. Reducing the number of shuffles to 1 degrades reward stability and can lead to slower convergence, indicating that multiple shuffles are important for learning a robust ordinal signal. We fix the window length to 15 and reward frequency to 15.

More detailed results of hyperparameters sweep can be found in~\ref{table:ablations_all}.

\begin{table*}[h]
\caption{\textbf{Ablation study on window length $M$, reward frequency, and number of shuffles $L$ on Catrap L2.} {\color{black}RTA achieves strong performance across a range of settings, with best results at intermediate values. Performance degrades slightly and variance increases at extreme configurations, indicating sensitivity to the choice of hyperparameters.}}
\centering
\begin{tabular}{lcc}
\toprule
\textbf{Ablation: Window Length $M$} & Success rate & Update steps \\
\midrule
5  & 0.73 ± 0.09 & 65.33 ± 12.76 \\
15 & \textbf{1.00 ± 0.00} & \textbf{37.00 ± 10.98} \\
25 & \textbf{1.00 ± 0.00} & 43.33 ± 5.79 \\
\midrule
\textbf{Ablation: Reward Frequency} & & \\
\midrule
5  & \textbf{1.00 ± 0.00} & 175.60 ± 76.53 \\
15 & \textbf{1.00 ± 0.00} & \textbf{37.00 ± 10.98} \\
25 & \textbf{1.00 ± 0.00} & 128.67 ± 84.32 \\
\midrule
\textbf{Ablation: Number of Shuffles $L$} & & \\
\midrule
1 & \textbf{1.00 ± 0.00} & 143.66 ± 63.43 \\
2 & \textbf{1.00 ± 0.00} & \textbf{37.00 ± 10.98} \\
4 & 0.93 ± 0.09 & 123.33 ± 83.83 \\
\bottomrule
\end{tabular}

\label{table:ablations_all}
\end{table*}

\subsection{Examples}
\label{appendix:examples}

\begin{figure*}
  \centering
  \includegraphics[width=0.7\linewidth]{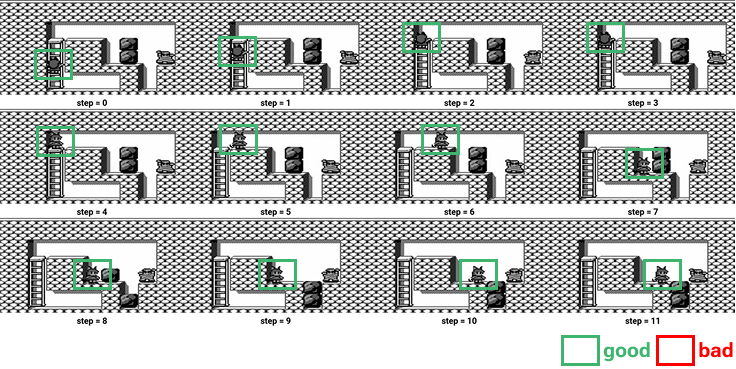}
  \caption{Example of good level completion with well-trained stage 1 model  with reward 0.77. {\color{black}The character (highlighted by a box) moves toward completing the level and receives a high reward from the scorer.}}
  \label{fig:good_welltrained}
\end{figure*}

\begin{figure*}
  \centering
  \includegraphics[width=0.7\linewidth]{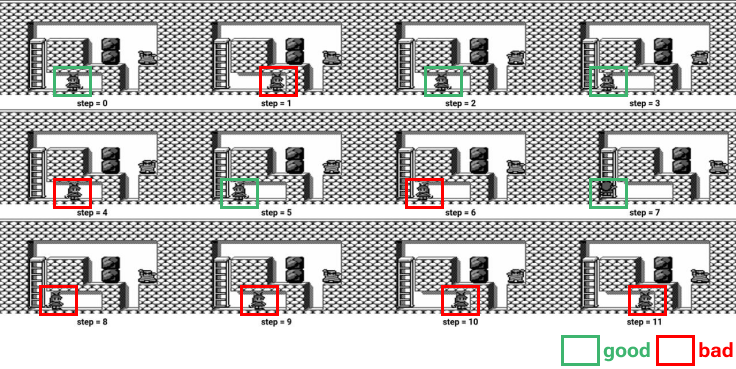}
  \caption{Example of bad level compeletion with well-trained stage 1 model  with reward -0.04. {\color{black} The character (highlighted by a box) moves back and forth without making progress, and is therefore assigned a low score.}}
  \label{fig:bad_welltrained}
\end{figure*}

\begin{figure*}
  \centering
  \includegraphics[width=0.7\linewidth]{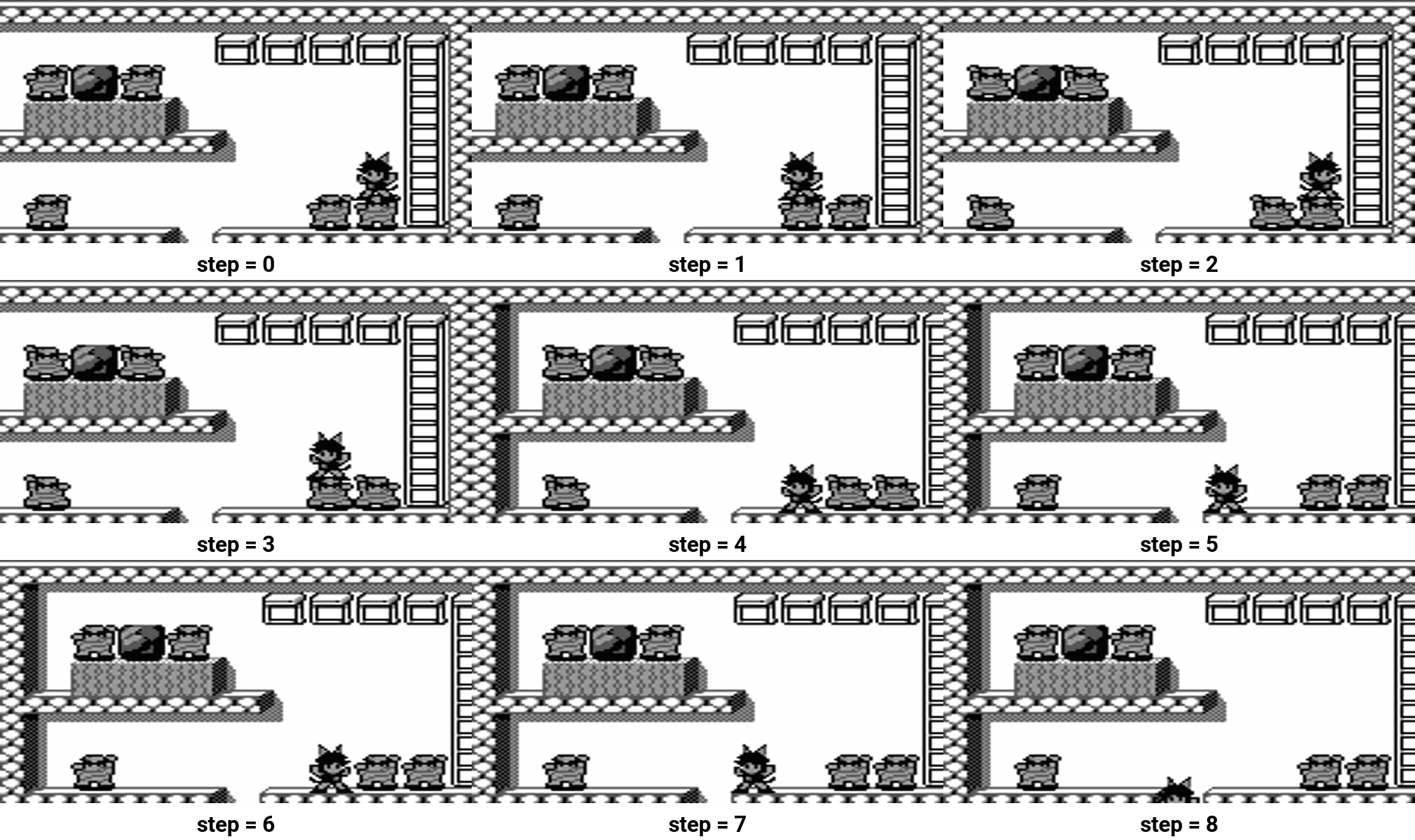}
  \caption{Example of bad level compeletion with undertrained stage 1 model  with reward 0.66 (high). {\color{black} We apply a scorer trained on a different level to Level~3. It still assigns a high score to a trajectory in which the character moves left but ends in a dead end.}}
  \label{fig:bad_undertrained}
\end{figure*}

\begin{figure*}
  \centering
  \includegraphics[width=0.7\linewidth]{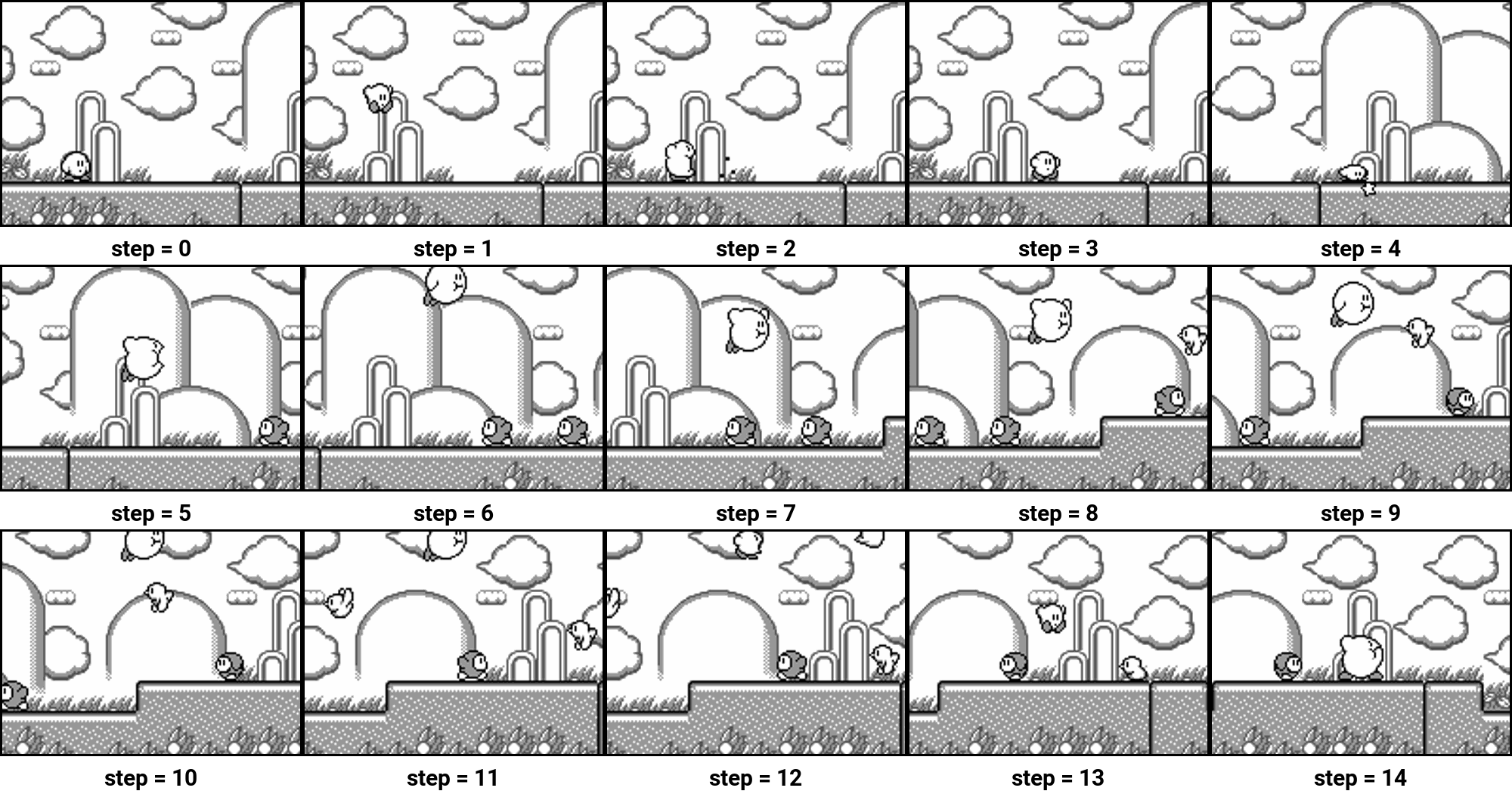}
  \caption{Example of good level compeletion with undertrained stage 1 model  with reward 0.15 (low). {\color{black} We apply a scorer trained on a different level to Kirby lvl~0. As this environment is out-of-distribution and includes a moving background, the scorer fails to assign a high score even when the character makes clear progress.}}
  \label{fig:good_undertrained}
\end{figure*}

\begin{figure}[t]
  \centering
  \includegraphics[width=0.8\linewidth]{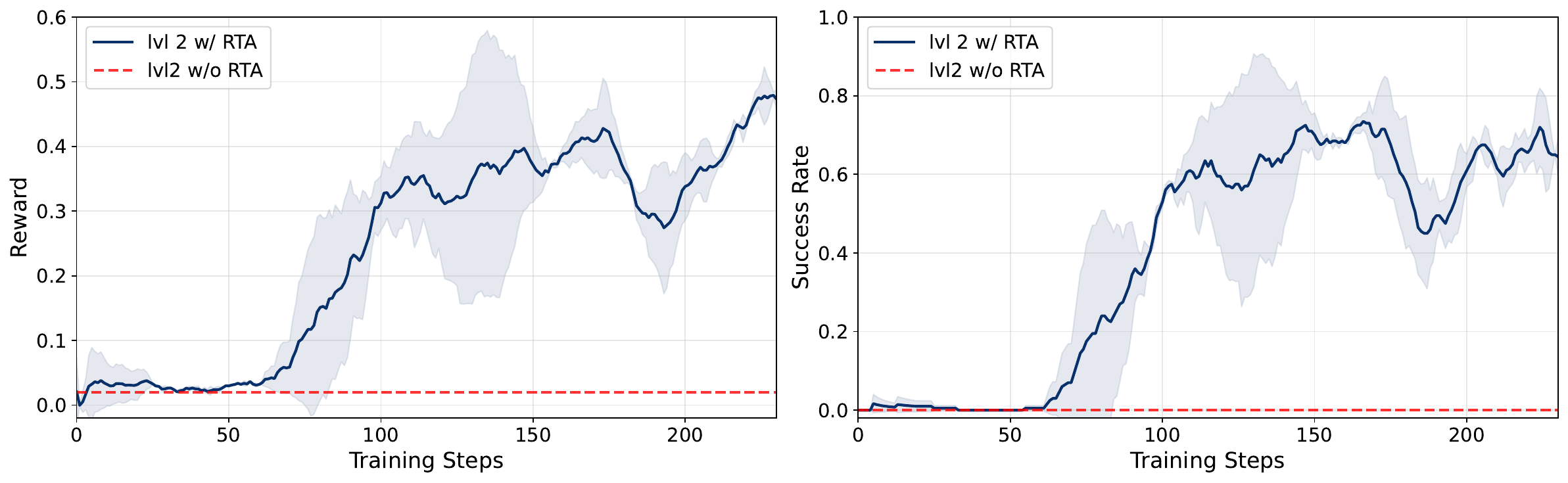}
  \caption{\textbf{Correlation-only reward drives control.} For Catrap L2 mean accumulated VOC-reward during training highly correlates with the success rate over episode.}
  \label{fig:stage2_reward_success}
\end{figure}

\textbf{Level 2 training curves.}
\label{appendix:lvl2_curve}
Figure~\ref{fig:stage2_reward_success} reports the \emph{success rate} (rolling window over episodes) alongside the per-query \emph{progress–time} reward when optimizing with VL-DAC~\citep{bredis2025enhancing}. In the plot, we show results from two random seeds, with quality smoothed with an EMA over the last 10 \emph{queries}. At several points during training, the RTA-trained agent attains $100\%$ success (if we turn off smoothing).

Although the reward is computed sparsely (every $N=15$ steps) and is always bounded in $[-1, 1]$, it provides sufficiently shaped feedback for policy learning: the mean reward increases steadily throughout training, closely tracking the agent's success rate.

This experiment demonstrates that the progress–time correlation signal-derived solely from video-based ordinal progress-is both informative and stable enough to drive effective online RL, even in the complete absence of extrinsic rewards.

Overall Pearson correlation for GameBoy games is reported in Table~\ref{table:reward_correlation}.

\begin{table*}[h]
\caption{Pearson correlation of mean cumulative reward and pass@5 during training. We report correlation of 0 if method is unable to solve level and N/A if experiment was not conducted. Results in this table reported across 5 seeds. {\color{black} RTA-assigned rewards exhibit a strong correlation with successful level completion. }}
\centering
\begin{tabular}{lcccc}
    \toprule
    Task        & Level 2  & Level 4 & Level 6 & Kirby (level 0) \\
    \midrule
    GVL   & -0.01 & 0.00 & -0.01 & 0.00  \\
    {\color{black}GVL-Gemini} & 0.33 &	0.00 & 0.00 & 0.00\\ 
    Rank2Reward              & -0.04 & -0.02 & -0.04 & N/A \\
    VLM-RM                   & 0.53 & -0.19 & -0.33 & 0.00 \\
    VLM-RM$_{\text{reg}}$ ($\alpha=0.5$) & 0.25 & 0.00 & -0.16 & 0.00 \\
    \midrule
    RTA (w stage1 training)    & \textbf{0.76} & \textbf{0.87} & \textbf{0.42} & \textbf{0.13} \\
    \bottomrule
\end{tabular}

\label{table:reward_correlation}
\vspace{-10pt}
\end{table*}

\begin{figure}[t]
  \centering
  \includegraphics[width=0.5\linewidth]{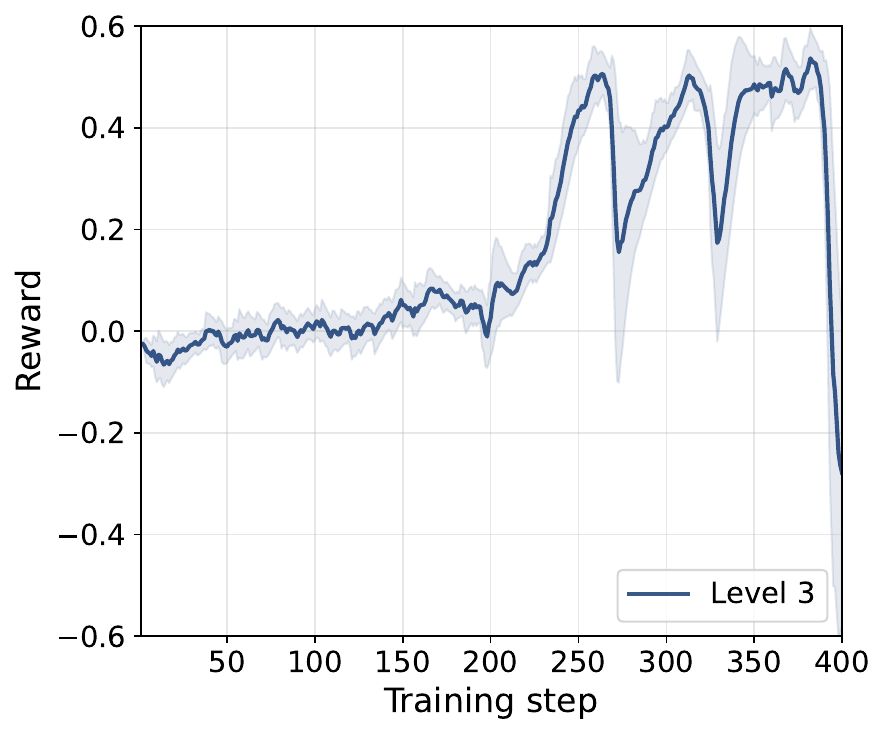}
  \caption{\textbf{Level 3 exposes training sensitivity.} Over 400 GRPO steps, \emph{progress–time} $\rho$ alternates between gains and dips, indicating that ambiguous content learns but benefits from careful data prep and sufficient duration.}
  \label{fig:stage1_level3_voc}
\end{figure}

\textbf{Level 3 performance}
\label{appendix:stage1_level3}
Notably, level 3 did not converge within 200 steps, prompting further experiments with longer training on this level (see Fig.~\ref{fig:stage1_level3_voc}).

\subsection{Training efficiency}
\label{app:efficiency}
\begin{figure}[t]
  \centering
  \includegraphics[width=0.8\linewidth]{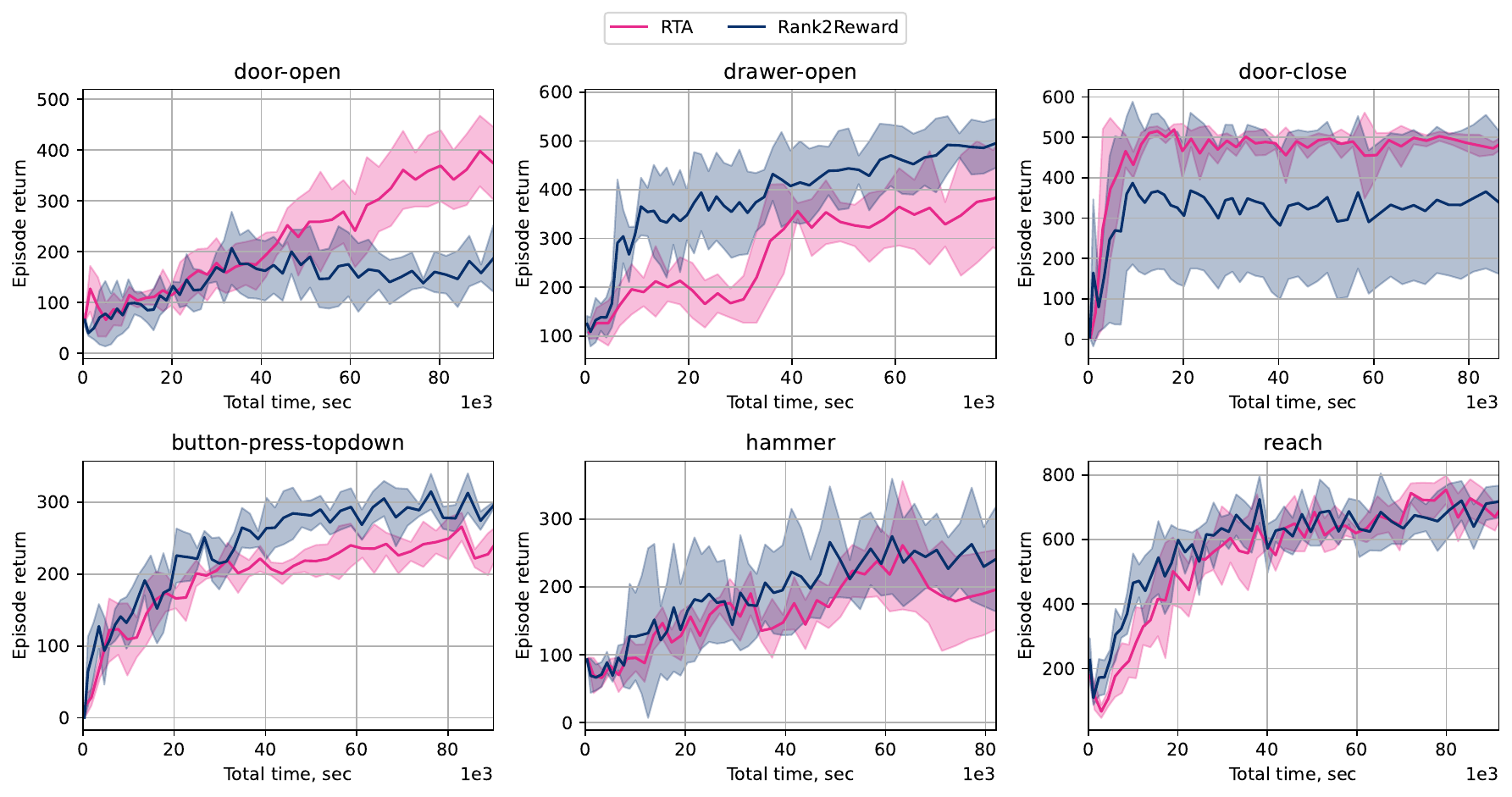}
  \caption{Training efficiency of RTA compared to Rank2Reward on MetaWorld tasks under aligned computational budget.}
  \label{fig:metaworld_compute}
\end{figure}

\begin{figure}[t]
  \centering
  \includegraphics[width=0.6\linewidth]{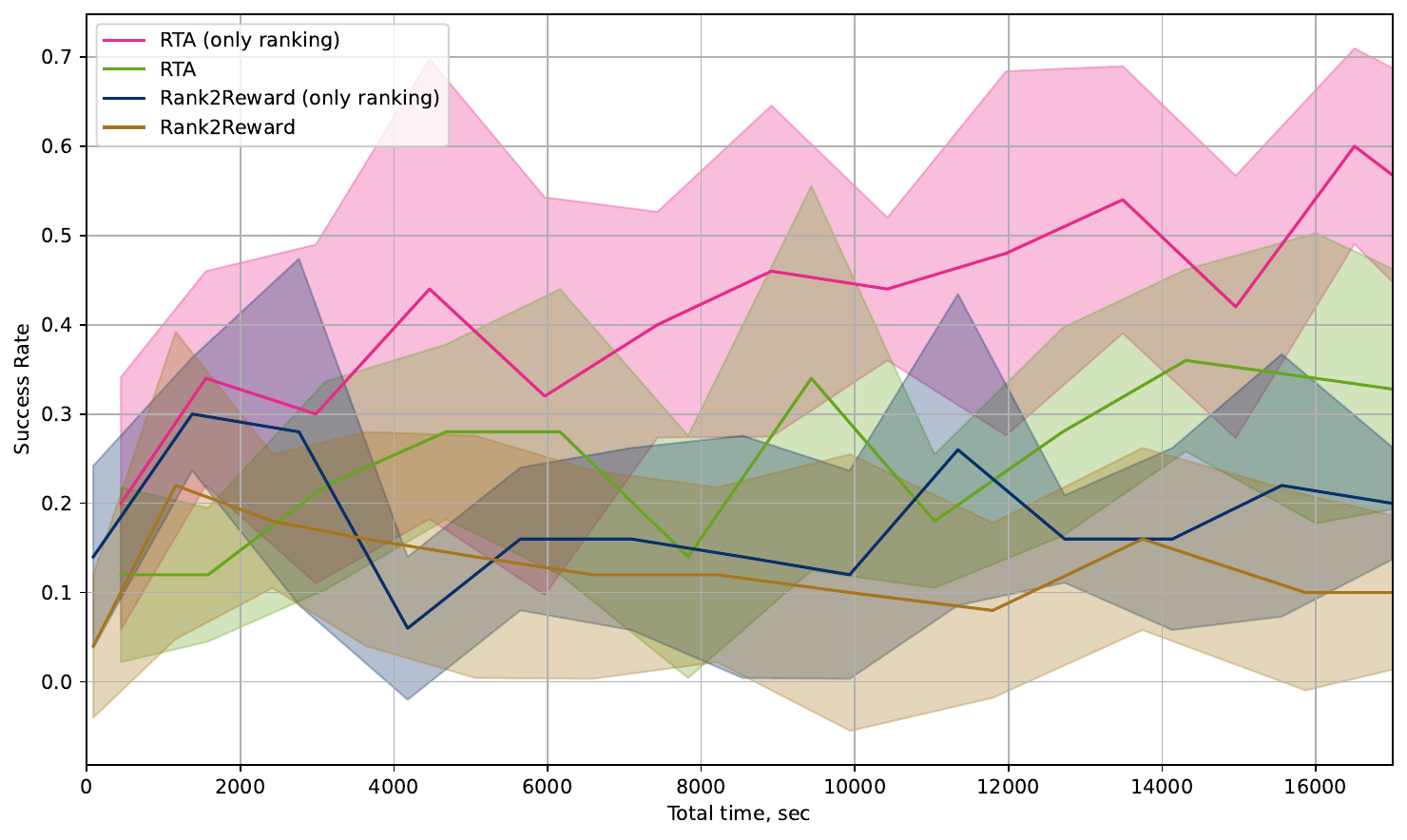}
  \caption{Training efficiency of RTA compared to Rank2Reward on UMaze under aligned computational budget.}
  \label{fig:umaze_compute}
\end{figure}

We have compared RTA and Rank2Reward under the same compute budget. Fig.~\ref{fig:metaworld_compute} reports episode return versus total wall-clock time, rather than training iterations, so the comparison is budget-aligned rather than step-aligned. Under this matched budget, RTA is stronger on \texttt{door-open}, \texttt{door-close}, and \texttt{reach}; it reaches near-parity on \texttt{drawer-open}; and Rank2Reward is somewhat stronger on \texttt{button-press-topdown} and \texttt{hammer}. The main point is therefore not that RTA wins every task, but that its advantage is not explained by extra compute. Under matched budget, it is competitive across all tasks and better on a majority of them. The same holds for PointMaze environment, RTA outperforms Rank2Reward given equal compute budget Fig.~\ref{fig:umaze_compute}.

{\color{black} \subsection{Cyclic trajectory construction.}
\label{app:cyclic}
We evaluate reward behavior using cyclic trajectories constructed by combining forward and reversed segments of expert demonstrations~Fig.\ref{fig:loop}. We mark the cyclic region explicitly and plot cumulative normalized Spearman correlation reward over time. We test this setup on expert trajectories from MetaWorld \texttt{door-open} task and one UMaze expert video from PointMaze. We use Stage~1 scorers that were trained per those tasks respectively.

The trajectory is divided into non-overlapping windows of varying sizes (1, 3, 5, 10, 15 seconds and full video). From each window, we uniformly sample a fixed number of frames and compute rank correlation with respect to temporal order.

We observe three regimes. For small windows (e.g., 1 second), the reward increases steadily even within cyclic regions, indicating sensitivity to local ordering but not global progress. For intermediate windows (e.g., 3–5 seconds), the cumulative reward decreases or stabilizes, better reflecting the absence of true progress. For large windows ($\geq$ 5 seconds), the signal approaches zero due to averaging across inconsistent temporal segments. Since the PointMaze trajectory is longer than the MetaWorld one, the segments outside the cyclic boundary are more pronounced. As a result, the 5-second window yields the most stable behavior when evaluated relative to the cycle boundary onset.

After the cyclic region ends and the trajectory resumes normal forward progress, all window sizes produce a sharp increase in reward, correctly identifying goal completion. These results highlight the importance of window size as a trade-off between local sensitivity and global consistency.

\textbf{Window size selection.}
Based on the analysis above, we fix a single intermediate window size across all experiments and do not tune it per task. Additional ablations on window size (Catrap Level~2) are provided in Table~\ref{table:ablations_all}. Empirically, this choice falls within a stable regime where the reward signal avoids both local artifacts from very small windows and signal dilution from very large ones, while remaining consistent across environments.}



\end{document}